# Solving generalized grouping problems in cellular manufacturing systems using a network flow model

Md. Kutub Uddin
*Department of Mechanical Engineering, Khulna University of Engineering & Technology, Khulna, Bangladesh*
Md. Saiful Islam
*Department of Industrial Engineering and Management, Khulna University of Engineering & Technology, Khulna, Bangladesh*
Md Abrar Jahin
*Thomas Lord Department of Computer Science, Viterbi School of Engineering, University of Southern California, Los Angeles, California, USA*
Md. Saiful Islam Seam
*Department of Industrial Engineering and Management, Khulna University of Engineering & Technology, Khulna, Bangladesh, and*
M. F. Mridha
*Department of Computer Science, American International University-Bangladesh, Dhaka, Bangladesh*

## Abstract

**Purpose** – This study addresses the generalized grouping problem in cellular manufacturing systems (CMS) where parts may have multiple alternative process routes. The objective is to optimally form process route families by minimizing dissimilarity based on required machines, without pre-specifying the number of families and to subsequently form machine cells that maximize machine utilization.

**Design/methodology/approach** – The process route family formation problem is formulated as a unit-capacity minimum-cost network flow model, exploiting the structure and efficiency of network flow algorithms. This constitutes the first stage of a hierarchical solution framework. In the second stage, machine cell formation is addressed using two approaches: a quadratic assignment programming (QAP) formulation and a hierarchical heuristic procedure. Both approaches assign process route families and machines to a fixed number of cells.

**Findings** – Computational experiments on test problems demonstrate that the proposed network flow model effectively forms process route families optimally. Results further show that the QAP formulation and the heuristic procedure for machine cell formation produce identical solutions, indicating the effectiveness of the heuristic as a computationally efficient alternative.

**Research limitations/implications** – The study assumes deterministic process routes and machine requirements, which may not fully capture variability in real manufacturing environments. Future research may extend the model to incorporate stochastic processing times, dynamic routing or scalability to large industrial systems.

*Funding:* This research did not receive any specific grant from funding agencies in the public, commercial or not-for-profit sectors.

*Competing interests:* The authors declare no competing interests related to this research.

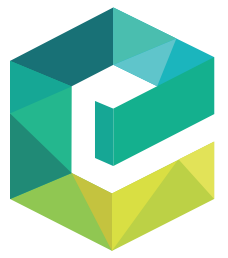

**Practical implications** – The proposed framework provides manufacturing planners with an exact and structured approach to forming process route families and machine cells. By eliminating the need to predefine the number of families and offering an efficient heuristic alternative, the method enhances practical applicability in CMS design.

**Social implications** – Improved efficiency in cellular manufacturing can contribute to reduced waste, better resource utilization and more sustainable industrial operations, indirectly supporting environmental and economic sustainability goals.

**Originality/value** – This research introduces a novel application of unit-capacity minimum-cost network flow modeling to process route family formation in generalized CMS. The integrated hierarchical framework, combining exact optimization and heuristic methods, offers both theoretical rigor and practical efficiency, extending existing CMS grouping methodologies.

## 1. Introduction

The manufacturing industry is experiencing rapid growth driven by increasing demand across diverse product categories. This growth is accompanied by transformative changes that are fundamentally reshaping manufacturing paradigms. Cellular manufacturing systems (CMS), based on the philosophy of group technology (GT), have been recognized as a technological innovation in job shop or batch-type production systems to gain economic advantages similar to those of mass production systems. CMS is capable of producing small-to medium-sized batches of a wide variety of parts in a flow-line manner. The concept of CMS involves dividing the entire production system into smaller autonomous subsystems to improve shop floor control, material handling, tooling and scheduling. This approach leads to decreased setup times, in-process inventories and throughput times. To achieve this decomposition, it is necessary to identify subsets of parts with similar processing or design requirements so that each subset of parts can be processed by a single subsystem. In the context of GT, each such subsystem is termed a machine cell and each subset of parts is referred to as a part family. The identification of machine cells and part families is known as machine-component grouping and is the first step in the design of a CMS. This partitioning of machines and parts in a factory is achieved through the application of GT. The application of GT does not depend on the degree of automation in a factory and hence, GT can be applied at any level of automation, from manual production to fully automated systems.

The grouping problem varies depending on factors like the availability of information and the level of decision-making. In a simple grouping problem, each part has only one process plan and each operation of the process plan can be performed on only one machine. As a result, each part has only one process route. Here, the terms “process plan” and “process route” have different meanings. A process plan lists the operations required to complete a part, while a process route lists the machines to which the various operations are assigned. On the other hand, in a generalized grouping problem, each part has more than one process route. The objective of the grouping problem, in this case, is to identify a single process route for each part and determine the process route families and machine cells. The goal is to ensure that each machine cell is capable of handling at least one process route family.

The generalized grouping problem is a significant area of research within the field of network flow models, where the primary objective is to optimally assign groups under certain constraints. Since the pioneering work on grouping based on graph theory by Rajagopalan and Batra (1975), numerous graph-theoretic methodologies have been developed by researchers over the past few decades. These approaches have been thoroughly studied in the most recent literature, including applications in social networks and computer science (Majeed and Rauf, 2020). This problem is particularly important in contexts such as telecommunications, transportation and logistics, where efficient resource allocation is critical. The graph-theory-based methodologies can be classified into two main categories: graph decomposition (or partition) and network flow.

**Table 1.** Types of graphs for grouping

| Type of graph | Nodes | Arcs and weightage on arcs |
|---|---|---|
| Machine graph | Machines | Arcs represent the relationships between machines, indicating that there is movement of one or more parts between the machines, constituting the pair of nodes. The weight on the arcs represents one of the following: dissimilarity between the two machines, the total number of part movements between the two machines or similar metrics |
| Part graph | Parts | Arcs representing the relationship between parts indicate that one or more machines are common to the pair of parts constituting the nodes. The weight on the arcs represents some kind of dissimilarity or similarity between the two parts |
| Machine-part graph | Machines and parts | Arcs indicate the relationship between a machine and a part, showing that a certain part utilizes a specific machine. The weight on an arc may indicate how long the part takes to process on the machine or it could simply be assigned a value of 1 to show that the part uses the machine. No arcs exist between two parts or between two machines |

The graph decomposition method constructs a graph from the given parts' process routes. The nodes in these graphs correspond to either machines, parts or both. The arcs represent the associations between nodes and carry a weight representing some kind of dissimilarity or interaction between the nodes. Depending on the node type, three types of graphs can be constructed: machine graph, part graph and machine-and-part graph (Werner, 2020). These are described in Table 1.

To obtain machine cells and part families, graphs are successively decomposed into sub-graphs, ensuring minimal interaction among them or resulting in disconnected sub-graphs. Numerous studies have shown that there is no single strategy that works most effectively for obtaining part families and machine cells. For example, some research (Uddin and Shanker, 2002; Prafulla and Kulkarni, 2021) discussed the use of genetic algorithm (GA) to solve generalized grouping problems in CMS, demonstrating improvements in optimizing these groupings without pre-specifying the number of groups. In the network flow method, a network is constructed by creating two nodes for each machine or part. The nodes are connected by directed arcs and the weight on the arcs represents some kind of cost, indicating dissimilarity between the parts or machines. These formulations solve the grouping problem optimally without pre-specifying the number of groups. A pioneering work by Lee and Garcia-Diaz (1993) formulated the machine grouping problem as a capacitated circulation network for a simple grouping case.

Network flow algorithms have been successfully applied to various production planning problems, including the simple grouping problem. However, no work has been reported on solving the generalized grouping problem using network flow algorithms. In this paper, a methodology is presented that employs a hierarchical approach to solve the generalized grouping problem when process routes are given for each part. In the first stage, process route families are formed using a unit capacity minimum cost network flow model. In the second stage, a heuristic procedure and a quadratic assignment model are presented for machine cell formation and the assignment of process route families to these cells.

The remainder of the paper is structured as follows. Section 2 reviews the relevant research on grouping problems and network flow models. The problem environment considered for the model creation is explained in Section 3. The solution strategy is described in Section 4. The proposed network model for the development of process route (or part) families is presented in Section 5. Section 6 describes the development of machine cells and the assignment of part families to these cells. The application of the proposed model to a few numerical issues is demonstrated in Section 7. Finally, Section 8 presents the conclusions and recommendations for future research.

## 2. Literature review

Extensive studies have been conducted based on the framework of network flow models, particularly in routing (Yan and Wong, 2009; Cova and Johnson, 2003), planning and scheduling (Rietz *et al.*, 2016), production optimization (Lerlertpakdee *et al.*, 2014), supply chain management and intelligent transportation (Hsu and Wallace, 2007; Rudi *et al.*, 2016), inventory management and distribution (Hovav and Tsadikovich, 2015), social network analysis (Gomez *et al.*, 2013), resilience assessment (Goldbeck *et al.*, 2019; Yin *et al.*, 2022), infrastructural interdependencies (Holden *et al.*, 2013), water supply planning (Hsu and Cheng, 2002), strategic mine planning (Topal and Ramazan, 2012) and energy systems (Quelhas *et al.*, 2007; Quelhas and McCalley, 2007). A number of graph theory-based studies on grouping problems have been published in the literature, where machines or parts are represented by the vertices of graphs and appropriately determined similarity coefficients are defined by the weights of the arcs. The grouping problem was first solved using graph theory by Rajagopalan and Batra (1975). Using the route cards of the parts, they developed a machine graph in which the vertices represented the machines and the edges represented Jaccard's similarity coefficients. Cells were identified as groups of vertices that were highly connected to one another.

For studying the optimization of cell formation, various algorithmic techniques have been used. Among these techniques, the hybrid genetic algorithm (GA) has been widely employed for machine-part grouping in CMS. The formation of machine groups and part families is a complex task in cellular manufacturing, which seeks to combine the fast production rate of flow lines with the flexibility of job shops. The integration of local search heuristics with GA, as proposed by Tariq *et al.* (2009), has demonstrated encouraging results in quickly obtaining optimal solutions, indicating the effectiveness of hybrid techniques in addressing real-world manufacturing challenges. Mak *et al.* (2000) proposed a GA to optimize cell development in manufacturing systems. Their approach provided practical insights into algorithmic techniques and strategies for optimization by dynamically adjusting parameters to maximize efficiency and reduce handling costs. Similarly, Salehi and Moghaddam, (2010) offered complex techniques and frameworks for performance evaluation, analyzing the implementation of GA to address cell formation problems. These methods have been further developed by hybrid GA proposed by James *et al.* (2007), demonstrating efficient optimization in practical situations.

The grouping problem was first solved by Lee and Garcia-Diaz (1993) as a circulation network flow problem. The idea is to group machines into cells so that each family of parts can be processed in a single machine cell. To achieve this, a directed graph for the machines is created and its network flow problem is solved. The solution to the network flow problem consists of one complete loop and several sub-loops, each loop corresponding to a machine cell. Compared to the *p*-median model, this approach is reported to have excellent potential for providing computationally efficient and optimal solutions for simple grouping problems. Cheng *et al.* (2019) developed generalized grouping strategies in coded caching and Garg and Arora (2018) developed fuzzy soft-set decision-making frameworks, demonstrating the flexibility of grouping techniques. These methods integrate expert preferences and efficiently manage uncertainty, offering powerful tools for decision-making. The use of grouping concepts in decision theory and computational problems is further illustrated by the grouping functions proposed by Bustince *et al.* (2012) and the generalized group testing procedures proposed by Malinovsky (2019). These approaches enhance efficiency and scalability across various fields. This thorough analysis highlights the usefulness and potential of network flow models in addressing a range of grouping problems.

Recent advances in graph-based and knowledge-driven methodologies have further enriched the landscape of manufacturing system design. Knowledge graphs, which represent entities and their relationships as structured semantic networks, have emerged as a powerful paradigm for integrating heterogeneous manufacturing data and enabling intelligent decision support. Lu *et al.* (2025) proposed a temporal knowledge graph reasoning method that fuses dynamic event embeddings with Hawkes process-based prediction for predictive maintenance of electromechanical equipment, demonstrating how graph-based representations can capture evolving relationships in complex industrial systems. Similarly, Su *et al.* (2025) developed a multi-layer knowledge graph architecture for digital twin systems in manufacturing processes, integrating a concept layer, a model layer and a decision layer to enhance predictive analysis and anomaly detection in aero-engine blade production. These works highlight the growing importance of graph-based formalisms beyond classical network flow in addressing modern manufacturing challenges. Digital twin frameworks integrate physical and virtual system representations to facilitate simulation, monitoring and optimization (Iliuţă *et al.*, 2024; Liu *et al.*, 2023). More recently, Cognitive Digital Twins have been proposed as an evolution of traditional digital twins, incorporating learning, reasoning and adaptive capabilities to support autonomous decision-making in complex industrial systems.

It is worth noting that various graph-based models exist for representing and solving manufacturing system problems, including Bayesian networks (BNs), Petri nets and knowledge graphs. Bayesian networks are particularly suited for modeling probabilistic dependencies and reasoning under uncertainty, making them valuable for fault diagnosis and quality prediction in manufacturing. However, BNs are fundamentally designed for probabilistic inference and do not naturally accommodate the assignment and flow-conservation constraints inherent in grouping problems. Similarly, while Petri nets excel at modeling concurrent and sequential processes, they are primarily oriented towards process flow analysis rather than combinatorial grouping optimization. The network flow model adopted in this study is specifically designed to solve the minimum-cost assignment problem, where the objective is to minimize total dissimilarity among process routes within families. With clearly specified supply and demand nodes, capacity limitations, and cost functions, network flow models' mathematical structure immediately translates to the generalized grouping problem's combinatorial structure, allowing for optimal or nearly optimal solutions using well-known algorithms. This structural alignment makes the network flow formulation inherently more suitable for the grouping problem than alternative graph-based models.

Furthermore, the concept of cognitive digital twins (CDTs) has garnered significant attention in the context of Industry 5.0, where integrating human expertise with machine intelligence is paramount. Su *et al.* (2024) proposed a three-layer cognitive digital twin model that leverages knowledge graphs to incorporate workers' knowledge and experience into intelligent manufacturing processes, comprising an ontology layer, a knowledge layer and a cognitive layer for advanced analysis and model evolution. CDTs employ ontological representations and simulation models that share conceptual similarities with the graph-based formulations used in this study. Specifically, the network flow model developed here could serve as a foundational decision-support component within a CDT framework, where the process route family formation and machine cell assignment outputs could be integrated into a digital twin's virtual model for real-time simulation, what-if analysis and dynamic reconfiguration of manufacturing cells. Such integration would enhance the CDT's reasoning capabilities by providing optimal grouping solutions that can be re-evaluated as manufacturing conditions evolve, thereby supporting adaptive and human-centric manufacturing decision-making.

It is evident that network flow models have been successfully applied across various domains, including social networks, decision-making, industry and infrastructure, to ensure computational efficiency and practical application. These models provide robust frameworks for enhancing resilience, improving decision-making and optimizing complex systems. Consequently, they offer valuable insights for the development of a network flow model for the generalized grouping problem.

## 3. Problem environment

For the proposed network flow model, a generalized grouping problem is considered where each part has more than one process route. The input to the problem is a two-dimensional (0–1) matrix showing the requirement of machines for various operations of each process route. The entry value in the matrix is 1 if an operation of a process route requires a particular machine and 0 otherwise. The grouping problem involves selecting one process route for each part from the given alternatives and grouping them into process route families, each of which can be processed by a single machine cell. Several process route families can be processed by a machine cell. The objective is to minimize the distance (dissimilarity) among the process routes within a family. The proposed model is expected to result in machine cells with the minimum possible inter-cell movements.

We consider a problem situation where there are $K$ parts and each part has a set of distinct process routes. Different process routes for a part generally require different sets of machines to complete the operations. In total, combining all the parts, there are $N$ process routes. There are M machines of different types and a maximum of $C$ cells are to be created. The objective is to minimize the inter-cell movements of parts and to maximize machine utilization.

It is important to note that the problem formulation assumes deterministic process routes, where each part's set of feasible process routes and machine requirements are known with certainty. This assumption is adopted to establish the mathematical foundations of the network flow model and to demonstrate its optimality properties. The implications of relaxing this assumption to accommodate stochastic manufacturing environments are discussed in Section 8.

## 4. The solution approach

The proposed model solves the problem in a hierarchical manner. First, process route families are formed based on minimum dissimilarity among the members of each family using a network flow model. In the second stage, machine cells are formed and process route families are assigned to these cells simultaneously, with the objective of maximizing machine utilization. For machine cell formation and process route family assignment, a heuristic and a QAP formulation are proposed.

The mechanism of process route family formation is as follows. First of all, for each part, there is a source or supply node ($k_s$) and a sink or demand node ($k_d$). For each process route, there are two nodes, $a$ and $b$ *and* a directed arc ($a$, $b$). From each part source node, arcs go to the corresponding process route node $a$. All arcs coming to a part sink node come from the corresponding process route node $b$. It is obvious that only one process route must be selected for each part. Let part $k$ have *TPR(k)* process routes. So, $TPR(k)-1$ routes are eliminated from consideration by including the corresponding arcs in directed paths from source to sink ($k_s \rightarrow i_a \rightarrow i_b \rightarrow k_d$). The flow value on all process route arcs is constrained to 1. The supply at the source and demand at the sink are fixed at $TPR(k)-1$ for all parts. This results in exactly one process route, with no supply from the source for each part. However, since all process routes are constrained to have flow value 1, even the process route arcs getting no supply from the source node have to have the flow conditions satisfied. This is possible only if such unsatisfied arcs form cycles within themselves, which are nothing but process route families. Further, to encourage better formation of process route cycles, the arc costs from one process route to another are taken as the corresponding dissimilarity values between them, the total of which is to be minimized over the whole network. In this way, the network flow model tries to group similar process routes together. The calculations of dissimilarity values are described in Sub-section 5.1.

At the stage of machine cell formation and route family assignment, the machine requirements for the various process route families are evaluated. Machine cells are created based on these requirements. Some process route families may be assigned to a cell if their machine requirements are fully met by that cell. If the requirements do not match completely, the family will be assigned to a cell where most of its requirements are satisfied, which may result in inter-cell movements.

The following notations are used for the development of the network flow model for process route family formation and machine cell formation model.

**Indices**

$i, j$ = process route
$k$ = part
$m$ = machine
$c$ = machine cell
$r$ = process route family

**Parameters**

$K$ = total number of parts
$PR(k)$ = set of process routes for part $k$
$TPR(k)$ = total number of process routes for part $k$
$= |PR(k)|$
$A$ = set of process routes over all parts
$N$ = total number of process routes over all parts
$= |A|$
$= \sum_{k=1}^{K} TPR(k)$
$M$ = total number of machines
$C$ = maximum number of machine cells that can be formed
$R$ = total number of process route families
$max_c$ = maximum number of machines that can be assigned to a cell $c$
$\omega(i,j)$ = cost of unit flow on arc $(i, j)$

$$a_{im} = \begin{cases} 1 & \text{if process route } i \text{ requires machine } m \text{ for processing} \\ 0 & \text{(irrespective of thenumber of times itis required} \\ & \text{otherwise} \end{cases}$$

$u_{mr}$ = usage factor for machine $m$ in process route family $r$ indicating the number of processes routes in a process route family $r$ using machine $m$
$= \sum_{i \in r} a_{im}$

**Decision variables**

$$Z_{mc} = \begin{cases} 1 & \text{if machine } m \text{ is assigned to cell } c \\ 0 & \text{otherwise} \end{cases}$$

$$Y_{rc} = \begin{cases} 1 & \text{if process route family } r \text{ is assigned to machine cell } c \\ 0 & \text{otherwise} \end{cases}$$

$f(x, y)$ = flow from node $x$ to node $y$

## 5. Proposed network flow model for route family formation

The proposed network flow model consists of three steps. The first step involves computing the pairwise dissimilarity between process routes, as described in subsection 5.1. The second step constructs the unit capacity minimum cost flow network, detailed in subsection 5.2. The third step identifies the process route families by solving the unit capacity minimum cost network flow problem, as described in subsection 5.3.

### *5.1 Computation of dissimilarities (or distances)*

The pairwise dissimilarities between process routes are computed as follows:

The dissimilarity value between a pair of process routes is measured as the number of machines that are not common to them. Let $d_{ij}$ be the dissimilarity between process routes $i$ and $j$. The value of $d_{ij}$ (dissimilarity) is an indicator showing the degree of dissimilarity between process routes $i$ and $j$. The value gets smaller as the two process routes require more and more common machines for processing. The process route-machine matrix $A = [a_{im}]$ $(i = 1, \ldots, N;$

$m = 1, \ldots, M$) is used to calculate the elements $d_{ij}$; of the dissimilarity matrix $D = [d_{ij}]$. The Hamming metric, to calculate the distance between a pair of binary row vectors, is used to calculate the dissimilarity between two process routes of different parts as shown below:

$$d_{ij} = \sum_{m=1}^{M} \delta\left(a_{im}, a_{jm}\right) \quad \forall i \in PR(k_1), j \in PR(k_2), k_1, k_2 = 1, \ldots, K, k_1 \neq k_2$$

$$\text{where } \delta\left(a_{im}, a_{jm}\right) = \begin{cases} 1 & if\ a_{im} \neq a_{jm} \\ 0 & otherwise \end{cases} \tag{1}$$

For example, consider process routes 1 and 3, where the number of machines in route 1 is 7 and the number of machines for route 3 is 5, out of which 3 machines are common. Then the dissimilarity between the pair (1, 3) is taken as (total number of machines – total number of common machines) = (5 + 7) – (2 × 3) = 6. To illustrate the calculation of the total dissimilarity value for a process route family, let us consider a process route family having process routes 1, 2, 3 and 4. Now, one cyclic order for this family could be (1, 2), (2, 3), (3, 4) and (4, 1). The dissimilarity value for this cyclic order will be $d_{12} + d_{23} + d_{34} + d_{41}$, where $d_{ij}$ is the dissimilarity value between processes routes $i$ and $j$. There will be 4! such cyclic orders ($n!$ cyclic orders for $n$ process routes) and some of them will have the same dissimilarity value because of the symmetric nature of the dissimilarity coefficients, $d_{ij} = d_{ji}$. The dissimilarity value for a process route family is taken as the minimum of total dissimilarity values amongst all cyclic orders.

*5.2 Construction of the network*

The construction of the unit capacity minimum cost flow network is based on the work of Lee and Garcia-Diaz (1993), who proposed the network flow formulation for a simple grouping problem. The network is constructed using the $N \times N$ dissimilarity matrix D obtained in subsection 5.1. The creation of nodes and arcs connecting the nodes and the assignment of weightage on arcs are carried out as follows:

*5.2.1 Creation of nodes.* 5.2.1.1 Creation of supply nodes. Create $K$ supply nodes designated as $1_s, 2_s, \ldots, K_s$ and one node corresponding to each part (the subscript $s$ denotes that it is a supply node). The supply capacity of the node $k_s$ is taken as $TPR(k)-1$ units, i.e. one less than the total number of process routes of part $k$; $k = 1, \ldots, K$.

5.2.1.2 Creation of demand nodes. Create $K$ demand nodes designated as $1_d, 2_d, \ldots, K_d$, again one node corresponding to each part (the subscript $d$ denotes that it is a demand node). The demand requirement of node $k_d$ is again taken as $TPR(k)-1$ units; $k = 1, \ldots, K$.

5.2.1.3 Creation of intermediate nodes. Corresponding to each process route $i$, create a pair of nodes $i_a$ and $i_b$. The two sets of nodes are designated as $1_a, 2_a, \ldots, N_a$ and $1_b, 2_b, \ldots, N_b$; $i = 1, \ldots, N$. These $2N$ nodes are used as transshipment nodes, i.e. these nodes do not possess any supply capacity, nor do they have any demand.

*5.2.2 Creation of arcs.* Create arcs and assign a capacity-cost triplet [$U$, $L$, $C$] to each arc to indicate an upper bound on its flow ($U$), a lower bound on its flow (L) and the per unit cost of flow ($C$) respectively, according to the following rules.

*Rule I*:

For each supply node $k_s$, $k = 1, \ldots, K$ and the corresponding transshipment node $i_a$, $i \in PR(k)$; create supply arcs directed from $k_s$ to $i_a$, signifying that part $k$ uses process route $i$. For example, arc ($1_s$, $1_a$) indicates that part 1 uses process route 1 for its processing. Assign capacity-cost triplet [$U, L, C$] = [1, 0, 0] to all arcs ($k_s$, $i_a$).

*Rule II*:

For each node pair $i_a$ and $i_b$, $i = 1, \ldots, N$; create transshipment arcs directed from $i_a$ to $i_b$. The two nodes $i_a$ and $i_b$, connected by a directed arc, represent a process route. For example, arc ($1_a$, $1_b$) represents process route 1, arc ($2_a$, $2_b$) represents process route 2 and so on. Assign capacity-cost triplet [$U$, $L$, $C$] = [1, 1, 0] to all arcs ($i_a$, $i_b$).

*Rule III*:

For each node $i_b$, $i \in PR(k)$, $k = 1, \ldots, K$ and the corresponding node $k_d$, create demand arcs directed from $i_b$ to $k_d$. The directed arc ($i_b$, $k_d$) signifies that part $k$ uses process route $i$; for example, the directed arc ($1_b$, $1_d$ay) indicates that part 1 uses process route 1 for its processing. Assign capacity-cost triplet [$U$, $L$, $C$] = [1, 0, 0] to all arcs ($i_b$, $k_d$).

*Rule IV*:

For each node $i_b$ and $j_a$, $i \in PR(k)$, $j \notin PR(k)$, $k = 1, \ldots, K$; create relational arcs directed from $i_b$ to $j_a$ and assign capacity-cost triplet [$U$, $L$, $C$] = [1, 0, $d_{ij}$]. The cost of shipping one unit of flow from node $i_b$ to node $j_a$ is taken as the dissimilarity value between process routes $i$ and $j$.

The network constructed for the problem environment described in section 4, according to the steps described above, is shown in Figure 1.

5.2.2.1 Observations.

(1) The upper and the lower bound on flow on transshipment arcs ($i_a$, $i_b$), $i = 1, \ldots, N$ is 1 unit. Thus, in accordance with the flow conservation, a node $i_b$ can supply 1 unit of flow to either destination node $j_a$ or to demand node $k_d$, $i \in PR(k)$, $j \notin PR(k)$, $k = 1, \ldots, K$.

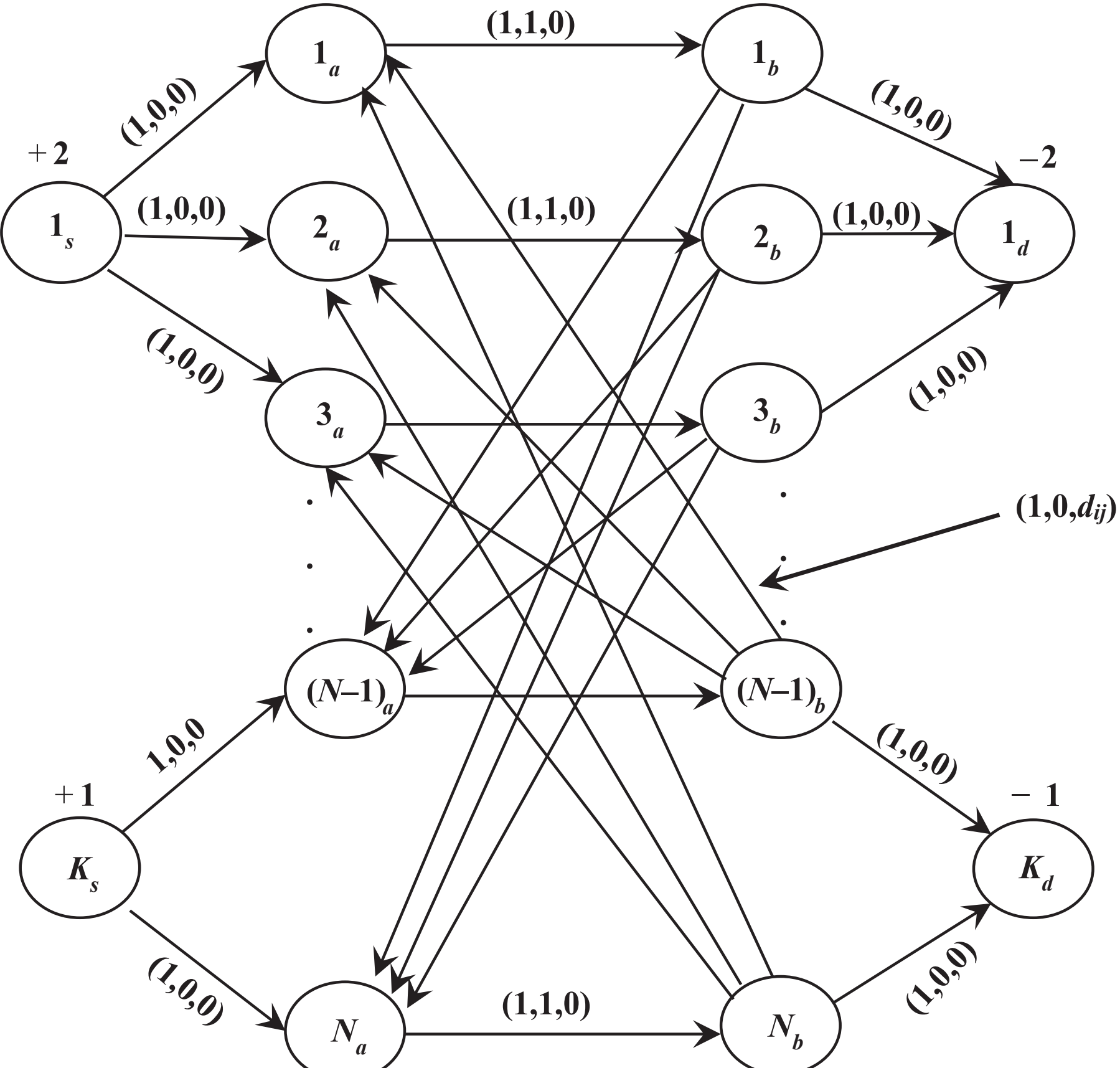

**Figure 1.** Network flow representation of process route family formation

(2) If a part $k$ has $TPR(k) = n$ process routes, $n$ units of flow are needed to satisfy the flow condition on the $n$ transshipment arcs, say, $(i_a, i_b)$, $((i+1)_a, (i+1)_b)$, . . . $((i+n-1)_a, (i+n-1)_b)$. However, according to the network construction, the supply node $k_s$ can supply only $TPR(k) - 1 = n - 1$ units of flow, that is, to only $n - 1$ of the transshipment arcs. The remaining 1 unit of flow required on the unsatisfied arc can be satisfied by including it either in a path with satisfied process route arcs of some other parts or in a cycle with unsatisfied process route arcs of some other parts. In general, such a cycle or path may involve more than one part and its process routes. The relational arcs $(i_b, j_a)$ ($i$ and $j$ must not belong to the same part) are used to create these cycles or paths and the weightage assigned to each such arc $(i_b, j_a)$ is the dissimilarity value between process routes $i$ and $j$, namely, $d_{ij}$. In the framework of the proposed minimum cost network flow problem, the relational arc that has the minimum dissimilarity (i.e. cost of flow) will be chosen for creating the cycle or path and this is true for all other parts. Therefore, any optimal (or feasible) solution of the proposed minimum cost network flow problem will contain paths (from supply nodes to demand nodes) with flow value 1 or both paths and cycles with flow value 1. Two types of paths could be there. The first kind, direct path ($k_s \rightarrow i_a \rightarrow i_b \rightarrow k_d$), includes only one process route, whereas the second kind, indirect path, includes more than one process route. The direct path does not contain any relational arcs, but the indirect path does.

(3) The cost of flow on all arcs is zero except on relational arcs $(i_b, j_a)$, $i \in PR(k), j \notin PR(k)$, $k = 1, \ldots, K$; whose cost of flow is the dissimilarity value between process routes $i$ and $j$, i.e. $d_{ij}$. Therefore, minimizing the total cost of flow in the network is equivalent to minimizing the total cost of flow on the relational arcs $(i_b, j_a)$ with flow value 1, which, in turn, is equivalent to minimizing the total dissimilarity values among the process routes involved in creating the paths and cycles.

*5.2.3 Mathematical formulation.* 5.2.3.1 Objective function. The objective function is to minimize the total cost of flow in the network and consists of four components: cost of flow on supply arcs, cost of flow on transshipment arcs, cost of flow on relational arcs and cost of flow on demand arcs. Assuming the linearity, these costs are computed as the product of flow rate $f$ (.,.) and cost of unit flow $\omega$ (.,.). The four cost components can be written as follows:

$$\text{Cost of flow on supply arcs} = \sum_{k=1}^{K} \sum_{i \in PR(k)} f(k_s, i_a).\omega(k_s, i_a)$$

$$\text{Cost of flow on transshipment arcs} = \sum_{i=1}^{N} f(i_a, i_b).\omega(i_a, i_b)$$

$$\text{Cost of flow on relational arcs} = \sum_{k=1}^{K} \sum_{i \in PR(k)} \sum_{j \notin PR(k)} f(i_b, j_a).\omega(i_b, j_a)$$

$$\text{Cost of flow on demand arcs} = \sum_{k=1}^{K} \sum_{i \in PR(k)} f(i_b, k_d).\omega(i_b, k_d)$$

The cost of flow on supply arcs, transshipment arcs and demand arcs will always be zero because their cost of flow is zero according to *Rule I*, *Rule II* and *Rule IV,* respectively. The cost of flow for the relational arcs, $\omega(i_b, j_a)$, is taken as $d_{ij}$, the dissimilarity value between the two process routes $i$ and $j$ according to *Rule III*. Therefore, the objective function is reduced to:

$$\min \sum_{k=1}^{K} \sum_{i \in PR(k)} \sum_{j \notin PR(k)} d_{ij} f(i_b, j_a) \tag{2}$$

5.2.3.2 Constraints.

(1) Flow conservation at source nodes $k_s$

Flow emanating from these nodes must be equal to their supply capacities:

$$\sum_{i \in PR(k)} f(k_s, i_a) = TPR(k) - 1 \quad k = 1, \ldots, K \tag{3}$$

(2) Flow conservation at nodes $i_a$

Since these are transshipment nodes, therefore, the incoming flow must be equal to the outgoing flow:

$$f(k_s, i_a) + \sum_{j=1}^{N} f(j_b, i_a) = 1 \quad k = 1, \ldots, K; \quad i \in PR(k); j \notin PR(k) \tag{4}$$

(3) Flow conservation at nodes $i_b$

These nodes are also transshipment nodes. Therefore, the incoming flow must be equal to the outgoing flow:

$$\sum_{j=1}^{N} f(i_b, j_a) + f(i_b, k_d) = 1 \quad k = 1, \ldots, K; \; i \in PR(k); \; j \notin PR(k) \tag{5}$$

(4) Flow conservation at sink nodes $k_d$

These are the sink nodes. Therefore, the flow incoming to these nodes must be equal to their requirement:

$$\sum_{i \in PR(k)} f(i_b, k_d) = -(TPR(k) - 1) \; k = 1, \ldots, K \tag{6}$$

(5) Side constraints - only one process route arc on any path from supply to demand node

This constraint is used to ensure that the path from a supply node to a demand node with flow value 1 must include only one process route and thus will eliminate from the network any indirect paths, i.e. the paths from supply nodes to demand nodes with flow value 1 which include more than one process routes. It can be written as:

$$f(k_s, i_a) = f(i_b, k_d) \; \forall i \in PR(k); \; k = 1, \ldots, K \tag{7}$$

(6) The integrality constraints for flow variables

$$f(k_s, i_a) = 0 \; or \; 1 \quad \forall i \in PR(k); \; k = 1, \ldots, K$$

$$f(i_b, k_d) = 0 \; or \; 1 \quad \forall i \in PR(k); \; k = 1, \ldots, K$$

$$f(i_b, j_a) = 0 \; or \; 1 \quad \forall i \in PR(k); \; j \notin PR(k); \; k = 1, \ldots, K \tag{8}$$

The model formulated in Equations 2–8 is a 0–1 integer linear programming problem, which can be solved using any procedure designed for 0–1 integer linear programming problems. Except for constraint 7, all other constraints and the objective function conform to the standard network structure and can be solved using any minimum-cost network flow algorithm. However, due to constraint 7, a specialized network flow algorithm is required to solve this problem. In this work, CPLEX, a general mixed integer optimization package (version 22.1.0.0), is used to solve the problem. The output consists of binary (0–1) flow values for all arcs.

### *5.3 Process route (part) family identification*

As stated in subsection 5.2 (observation 2), the network solution will contain cycles and paths (indirect paths) to satisfy the flow conditions on unsatisfied arcs. However, due to side constraint 7, the solution cannot contain any indirect paths, i.e. paths that include more than one process route and contain relational arcs. Therefore, the only way to satisfy the flow conditions on unsatisfied arcs is by creating cycles. The cost of flow on these arcs is the dissimilarity value between the two process routes connected by a relational arc. Since the objective function minimizes the cost of flow on relational arcs and only cycles can contain relational arcs, minimizing the total cost of flow on relational arcs is equivalent to minimizing the cost of flow on cycles. In other words, this means minimizing the dissimilarity values (which are the costs of flow for relational arcs) between the process routes included in the cycles. Generally, a cycle contains at least two process routes (each from a different part). These cycles will represent process route families in the context of GT and can be identified from the solution of the mathematical model described in subsection 5.2 by isolating the relational arcs with a flow value of 1. This identification of cycles is illustrated in Figure 2. The network shown in this figure represents a grouping problem with 4 machines and 4 parts (with a total of 8 process routes). To illustrate the identification of cycles, let the solution of the model have flow value 1 on relational arcs $(2_b, 3_a)$, $(3_b, 2_a)$, $(6_b, 7_a)$ and $(7_b, 6_a)$. From these, two cycles can be identified. The first cycle consists of the following four arcs:

$$\text{Cycle 1}: f(2_b, 3_a) = f(3_a, 3_b) = f(3_b, 2_a) = f(2_a, 2_b) = 1$$

The second cycle also consists of four arcs:

$$\text{Cycle 2}: f(6_b, 7_a) = f(7_a, 7_b) = f(7_b, 6_a) = f(6_a, 6_b) = 1$$

From these two cycles, two process route families can be identified:

(1) Process route family 1 consists of process routes 1 and 4, corresponding to parts 1 and 2, respectively.

(2) Process route family 2 consists of process routes 6 and 7, corresponding to parts 3 and 4, respectively.

### *5.4 Computational complexity and scalability analysis*

The computational complexity of the proposed network flow model is governed by the dimensions of the resulting 0–1 integer linear programming (ILP) problem. For a problem instance with K parts, $N$ total process routes across all parts and M machines, we characterize the model dimensions as follows. The network consists of $2K + 2N$ nodes: K supply nodes ($k_s$), K demand nodes ($k_d$) and 2N transshipment nodes ($i_a$ and $i_b$ for each process route). The number of arcs comprises four categories: N supply arcs ($k_s \rightarrow i_a$), N transshipment arcs ($i_a \rightarrow i_b$), N demand arcs ($i_b \rightarrow k_d$) and relational arcs ($i_b \rightarrow j_a$) where $i \in PR(k)$ and $j \notin PR(k)$. The number of

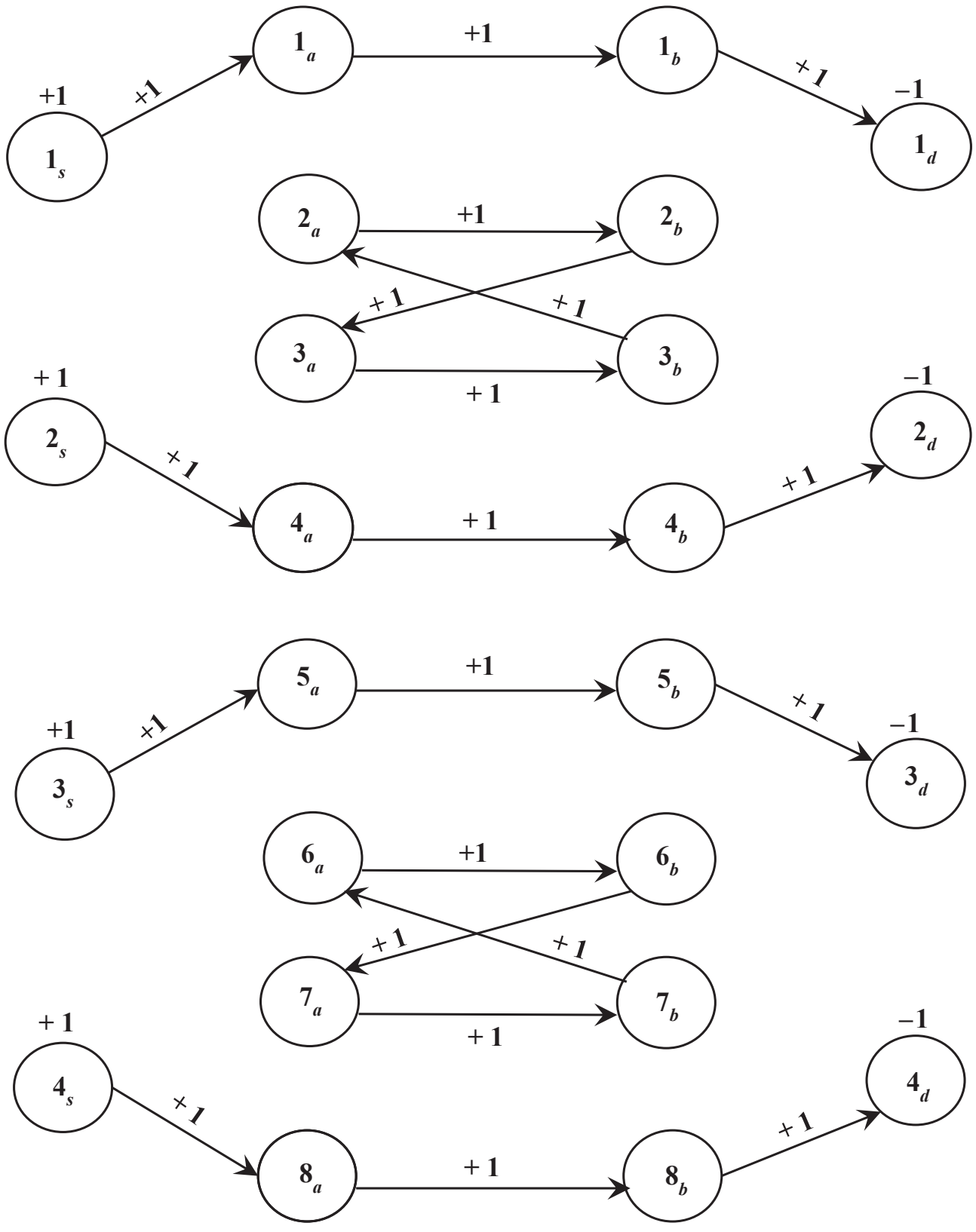

**Figure 2.** Network showing the paths and cycles for process route family formation

$$\text{Relational arcs} = \sum_{k=1}^{K} TPR(k)(N - TPR(k)) \tag{9}$$

$$\text{Upper bound:} \sum_{k=1}^{K} TPR(k)(N - TPR(k)) \leq N(N-1) \tag{10}$$

$$\text{Average case:} \sum_{k=1}^{K} TPR(k)(N - TPR(k)) = N(N - \overline{n}) \tag{11}$$

Where, $\overline{n} = \frac{N}{K}$ is the mean number of process routes per part. Each arc corresponds to a binary decision variable. Note that the $N$ transshipment arc variables are fixed at 1 (since the lower and upper bounds are both 1 by Rule II), leaving

$$\text{Free binary variables} = 2N + \sum_{k=1}^{K} TPR(k)(N - TPR(k)) \tag{12}$$

The total number of constraints comprises five groups: K supply-flow conservation constraints (Eq. 3), N flow conservation constraints at nodes $i_a$ (Eq. 4), N flow conservation constraints at nodes $i_b$ (Eq. 5), K demand-flow conservation constraints (Eq. 6) and N side constraints (Eq. 7), yielding a total of 3N + 2K explicit constraints (excluding integrality constraints, which are handled implicitly by the solver).

Table 2 provides the exact model dimensions and computational performance for the two example problems. All experiments were conducted using IBM ILOG CPLEX 22.1.0.0 on a desktop computer with an Intel Core i7 processor and 16 GB RAM, running under Windows 10.

From a theoretical perspective, the standard minimum-cost network flow problem (without side constraints) can be solved in strongly polynomial time. Well-known algorithms include the successive shortest path algorithm with complexity $O(n^2m)$, the minimum mean cycle-canceling algorithm with complexity $O(n^2m^3 \log n)$ and the network simplex method, where n and m denote the number of nodes and arcs, respectively (Ahuja *et al.*, 1993). Our formulation conforms to this standard structure in all respects except for side constraint (7), which couples the supply-side and demand-side flow variables for each process route, i.e. $f(ks, ia) = f(ib, kd)$. This side constraint introduces a structure analogous to the "network flow with side constraints" class of problems. While such problems are NP-hard in general, the integrality of our formulation (with unit capacities) and the specific structure of constraint (7) often allow CPLEX to solve them efficiently via LP relaxation with branch-and-bound, as the LP relaxation of network flow problems naturally yields integer solutions for the network-structured portion of the constraints.

Regarding scalability, the number of relational arcs (and thus binary variables) grows quadratically with N in the worst case. For a problem with K = 100 parts and an average of 5 process routes per part ($N = 500$), the number of relational arcs would be approximately $500 \times 495 = 247{,}500$, yielding a model with roughly 248,500 binary variables and 1,700 constraints. While modern commercial ILP solvers such as CPLEX and Gurobi routinely solve problems of this scale (and significantly larger), we acknowledge that instances with thousands of process routes may require computational times that are non-negligible. To address such cases, three mitigation strategies are available:

(1) LP relaxation exploitation: Since the constraint matrix is nearly totally unimodular (all constraints except Eq. (7) form a network matrix), the LP relaxation is expected to produce integer or near-integer solutions in most cases, significantly reducing the branch-and-bound search tree.

(2) Decomposition: The problem can be decomposed by partitioning parts into clusters based on shared machine requirements before solving smaller sub-problems independently. This approach trades global optimality for computational tractability while preserving solution quality in practice.

**Table 2.** Model dimensions and computational performance

| Characteristic | Example I | Example II |
|---|---|---|
| Number of parts (K) | 5 | 20 |
| Total process routes (N) | 11 | 51 |
| Number of machines (M) | 4 | 20 |
| Total nodes (2K + 2N) | 32 | 142 |
| Total arcs (binary variables) | 129 | 2,861 |
| —Supply arcs | 11 | 51 |
| —Transshipment arcs (fixed) | 11 | 51 |
| —Demand arcs | 11 | 51 |
| —Relational arcs | 96 | 2,708 |
| Total constraints (3N + 2K) | 43 | 193 |
| CPLEX solution time | <0.1 s | <0.5 s |
| Optimal objective value (min dissimilarity) | 0 | 4 |
| Exceptional elements (proposed model) | 0 | 1 |
| Exceptional elements (*p*-median model) | – | 3 |

(3) Heuristic alternative: The heuristic procedure proposed in Section 6.2 does not require an ILP solver and operates with time complexity dominated by O($R^2$ × M) for the merging steps, where $R$ is the number of process route families and M is the number of machines. This is substantially less than the ILP formulation. In our experiments, the heuristic yielded solutions identical to the exact ILP formulation for both example problems, suggesting it is a reliable alternative for large-scale instances.

It is also instructive to compare the computational requirements of the proposed model with those of alternative methods for the generalized grouping problem. Genetic algorithm-based approaches (Uddin and Shanker, 2002; Tariq *et al.*, 2009) typically require specifying population sizes, crossover/mutation rates and termination criteria and their convergence is not guaranteed to produce optimal solutions. The p-median model, used as a benchmark in Example II, requires pre-specification of the number of groups and solves a different optimization formulation. By contrast, the proposed network flow model provides provably optimal solutions to the process route family formation problem without pre-specifying the number of families and its solution time was negligible for both example problems tested.

## 6. Machine cell formation

For machine cell formation, we propose a QAP formulation and a heuristic procedure. The descriptions are given in the following subsections.

### *6.1 The QAP formulation*

The QAP formulation assigns part families and machines to cells simultaneously. The inputs to the model are the maximum number of cells that can be formed, the maximum number of machines that can be assigned to a cell and the machine usage factors (i.e. the number of process routes in a particular family using a particular machine) for all process route (part) families. The objective of the formulation is to maximize machine utilization. The output of the model is the assignment of part families and machines to cells.

*6.1.1 Objective function.* The objective of the model is to maximize machine utilization and can be written as:

$$\max \sum_{c=1}^{C} \sum_{r=1}^{R} \sum_{m=1}^{M} u_{mr} Z_{mc} Y_{rc} \tag{13}$$

*6.1.2 Constraints.* (1) Each machine is assigned to only one cell

$$\sum\nolimits_{c=1}^{C} Z_{mc} = 1 \quad m = 1, \ldots, M \tag{14}$$

(2) Each process route family is assigned to only one cell

$$\sum\nolimits_{c=1}^{C} Y_{rc} = 1 \quad r = 1, \ldots, R \tag{15}$$

(3) Maximum number of machines that can be assigned to a cell

$$\sum\nolimits_{m=1}^{M} Z_{mc} \le \max_{c} \quad c = 1, \ldots, C \tag{16}$$

*6.2 The heuristic procedure*
The objective of the heuristic procedure is also to maximize machine utilization. It follows a hierarchical approach and consists of three steps. The first step combines process route families wherever possible. The second step assigns machines to the process route families. The third step merges process route families and machine cells wherever feasible within the system constraints. The inputs to the procedure are the maximum number of machines that can be assigned to a cell, the process route families and the machines they require. The details of the heuristic are as follows:

*Step-1.* Combining the process route families

(1) Take a process route family $r$

(2) Check if the machines used by this process route family $r$ are a subset or superset of the machines used by any other process route family. If the answer is yes, merge the two process route families; otherwise, go to 3.

(3) Repeat 1 and 2 for all route families till no merging is possible.

*Step-2.* Assigning the machines to process route families

(1) Compute for each machine $m$ the usage factor $u_{mr}$ for each process route family $r$

(2) Assign machine $m$ to a process route family $r$ where $u_{mr}$ is maximum, the ties being broken arbitrarily.

(3) Repeat 2 for all machines.

*Step-3.* Merging the process route families and machine cells

Merge two process route families and the corresponding machine cells if there is some inter-cell movement between them, but only if the merging does not violate the cell size limit. Stop when no more merging is possible.

**7. Numerical illustrations**
To evaluate the effectiveness of the proposed formulation, two benchmark datasets are selected from the literature on generalized grouping problems: one without exceptional elements (i.e. where disjoint groups exist) and the other with exceptional elements (i.e. where disjoint groups do not exist). The number of exceptional elements serves as a critical performance metric for cell formation effectiveness The nusmber of exceptional elements is used as a critical criterion for measuring the effectiveness of cell formation, it is widely used by researchers in generalized grouping theory.

*7.1 Example problem I (without exceptional elements)*
The first problem shown in Table 3 is the incidence matrix. There are 5 parts with a total of 11 process routes and 4 machines. In the process route family formation stage, two cycles are identified.

$$\text{Cycle 1}: f(2_a, 2_b) = f(2_b, 7_a) = f(7_a, 7_b) = f(7_b, 2_a) = 1$$

$$\text{Cycle 2}: f(5_b, 11_a) = f(11_a, 11_b) = f(11_b, 9_a) = f(9_a, 9_b) = f(9_b, 5_a) = f(5_a, 5_b) = 1$$

The two cycles are shown in Figures 3 and 4, respectively. From the two cycles, two process route families can be identified.

(1) Process route family 1 consists of process routes 2 and 7, corresponding to parts 1 and 3, respectively.

**Table 3.** Example problem I

| Part (*k*) | Process route for part *k* ($i \in PR(k)$) | Process route SI. No. ($i \in A$) | Machine no. (*m*) 1 | 2 | 3 | 4 |
|---|---|---|---|---|---|---|
| 1 | 1 | 1 | 0 | 0 | 1 | 1 |
| | 2 | 2 | 0 | 1 | 0 | 1 |
| | 3 | 3 | 1 | 1 | 0 | 0 |
| 2 | 1 | 4 | 0 | 1 | 1 | 0 |
| | 2 | 5 | 1 | 0 | 1 | 0 |
| 3 | 1 | 6 | 1 | 0 | 0 | 1 |
| | 2 | 7 | 0 | 1 | 0 | 1 |
| 4 | 1 | 8 | 1 | 0 | 0 | 1 |
| | 2 | 9 | 1 | 0 | 1 | 0 |
| 5 | 1 | 10 | 0 | 0 | 1 | 1 |
| | 2 | 11 | 1 | 0 | 0 | 0 |

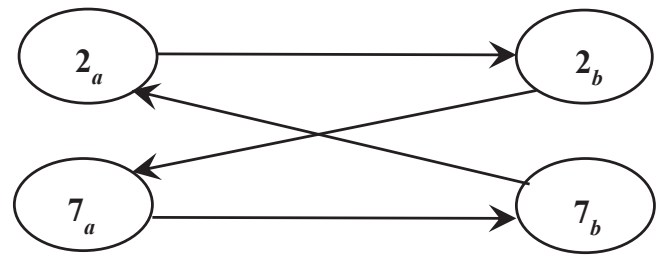

**Figure 3.** Cycle 1 for part family 1 of example problem I

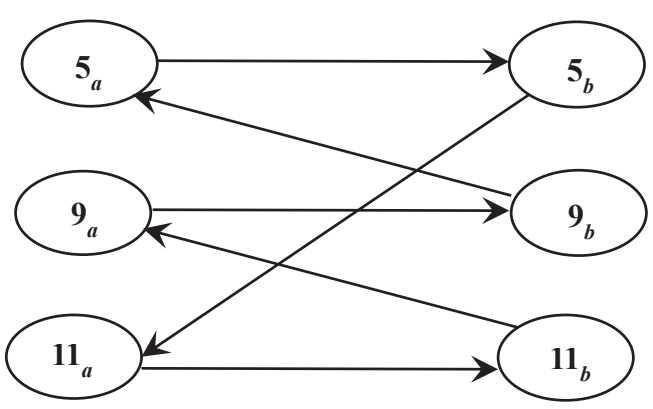

**Figure 4.** Cycle 2 for part family 2 of example problem I

(2) Process route family 2 consists of process routes 5, 9 and 11, corresponding to parts 2, 4 and 5, respectively.

In the machine cell formation stage, the machine utilizations $u_{mr}$ for each process route family $r$ are calculated and shown in Table 4.

From Tables 4 and it is obvious that none of the machines are utilized by more than one process route family and hence, two machine cells are easily identified as:

(1) Machine cell 1 consisting of machines 2 and 4

(2) Machine cell 2 consisting of machines 1 and 3

After rearranging the matrix according to the results obtained from the process route family formation and the machine cell formation stage, the resultant matrix is shown in Table 5.

**Table 4.** Machine utilization $u_{mr}$ for example problem I

| Process route family *(r)* → machine *(m)* ↓ | 1 | 2 |
|---|---|---|
| 1 | 0 | 3 |
| 2 | 2 | 0 |
| 3 | 0 | 2 |
| 4 | 2 | 0 |

**Table 5.** Solution to example problem I in block diagonal form

| Part *(k)* | Process route for part k *(i ∈ PR(k))* | Process route SI. No. *(i ∈ A)* | Machine no. (*m*) | | | |
|---|---|---|---|---|---|---|
| | | | 1 | 3 | 2 | 4 |
| 2 | 2 | 5 | 1 | 1 | 0 | 0 |
| 4 | 2 | 9 | 1 | 1 | 0 | 0 |
| 5 | 2 | 11 | 1 | 0 | 0 | 0 |
| 1 | 2 | 2 | 0 | 0 | 1 | 1 |
| 3 | 2 | 7 | 0 | 0 | 1 | 1 |

*7.2 Example problem II (with exceptional elements)*

The second problem is given in Table 6. This problem has 20 parts having a total of 51 process routes and 20 machines. In the process route family formation stage, seven cycles are identified as:

$$\text{Cycle 1}: f(2_a, 2_b) = f(2_b, 11_a) = f(11_a, 11_b) = f(11_b, 6_a) = f(6_a, 6_b) = f(6_b, 2_a) = 1$$

$$\text{Cycle 2}: f(30_a, 30_b) = f(30_b, 33_a) = f(33_a, 33_b) = f(33_b, 36_a) = f(36_a, 36_b) = f(36_b, 30_a) = 1$$

$$\text{Cycle 3}: f(49_a, 49_b) = f(49_b, 51_a) = f(51_a, 51_b) = f(51_b, 50_a) = f(50_a, 50_b) = f(50_b, 49_a) = 1$$

$$\text{Cycle 4}: f(4_a, 4_b) = f(4_b, 8_a) = f(8_a, 8_b) = f(8_b, 4_a) = 1$$

$$\text{Cycle 5}: f(41_a, 41_b) = f(41_b, 47_a) = f(47_a, 47_b) = f(47_b, 41_a) = 1$$

$$\text{Cycle 6}: f(44_a, 44_b) = f(44_b, 48_a) = f(48_a, 48_b) = f(48_b, 44_a) = 1$$

$$\text{Cycle 7}: f(12_a, 12_b) = f(12_b, 17_a) = f(17_a, 17_b) = f(17_b, 23_a) = f(23_a, 23_b) = f(23_b, 21_a) = f(21_a, 21_b) = f(21_b, 28_a) = f(28_a, 28_b) = f(28_b, 12_a) = 1$$

**Table 6.** Example problem II

| Part (*k*) | Process route set (*PR(k)*) | Process route number (*i*) | Machine no. (*m*) | | | | | | | | | | | | | | | | | | | |
|---|---|---|---|---|---|---|---|---|---|---|---|---|---|---|---|---|---|---|---|---|---|---|
| | | | 1 | 2 | 3 | 4 | 5 | 6 | 7 | 8 | 9 | 10 | 11 | 12 | 13 | 14 | 15 | 16 | 17 | 18 | 19 | 20 |
| 1 | 1 | 1 | 0 | 0 | 0 | 0 | 0 | 1 | 0 | 0 | 1 | 0 | 0 | 1 | 0 | 0 | 0 | 0 | 0 | 0 | 0 | 0 |
| | 2 | 2 | 0 | 0 | 0 | 0 | 0 | 0 | 1 | 0 | 1 | 0 | 0 | 1 | 0 | 0 | 0 | 0 | 0 | 0 | 0 | 0 |
| 2 | 1 | 3 | 1 | 0 | 0 | 0 | 0 | 1 | 0 | 0 | 0 | 0 | 0 | 1 | 0 | 0 | 0 | 0 | 0 | 0 | 0 | 0 |
| | 2 | 4 | 1 | 0 | 0 | 0 | 0 | 0 | 1 | 0 | 0 | 0 | 0 | 1 | 0 | 0 | 0 | 0 | 0 | 0 | 0 | 0 |
| 3 | 1 | 5 | 1 | 0 | 0 | 0 | 0 | 1 | 0 | 0 | 1 | 0 | 0 | 1 | 0 | 0 | 0 | 0 | 0 | 0 | 0 | 0 |
| | 2 | 6 | 1 | 0 | 0 | 0 | 0 | 0 | 1 | 0 | 1 | 0 | 0 | 1 | 0 | 0 | 0 | 0 | 0 | 0 | 0 | 0 |
| 4 | 1 | 7 | 1 | 0 | 0 | 0 | 0 | 1 | 0 | 0 | 0 | 0 | 0 | 1 | 0 | 0 | 0 | 0 | 0 | 0 | 0 | 0 |
| | 2 | 8 | 1 | 0 | 0 | 0 | 0 | 0 | 1 | 0 | 0 | 0 | 0 | 1 | 0 | 0 | 0 | 0 | 0 | 0 | 0 | 0 |
| 5 | 1 | 9 | 1 | 0 | 0 | 0 | 0 | 0 | 0 | 0 | 1 | 0 | 0 | 1 | 0 | 0 | 0 | 0 | 0 | 0 | 1 | 0 |
| | 2 | 10 | 1 | 0 | 0 | 0 | 0 | 0 | 0 | 0 | 1 | 0 | 0 | 1 | 0 | 0 | 0 | 0 | 0 | 1 | 0 | 0 |
| | 3 | 11 | 1 | 0 | 0 | 0 | 0 | 0 | 0 | 0 | 1 | 0 | 0 | 1 | 0 | 0 | 0 | 0 | 0 | 0 | 0 | 1 |
| 6 | 1 | 12 | 0 | 1 | 0 | 0 | 1 | 1 | 0 | 0 | 0 | 0 | 0 | 0 | 0 | 0 | 0 | 0 | 0 | 0 | 0 | 0 |
| | 2 | 13 | 0 | 1 | 0 | 0 | 1 | 0 | 1 | 0 | 0 | 0 | 0 | 0 | 0 | 0 | 0 | 0 | 0 | 0 | 0 | 0 |
| 7 | 1 | 14 | 0 | 0 | 0 | 0 | 1 | 0 | 1 | 0 | 0 | 0 | 0 | 0 | 0 | 0 | 0 | 1 | 0 | 0 | 1 | 0 |
| | 2 | 15 | 0 | 0 | 0 | 0 | 1 | 0 | 1 | 0 | 0 | 0 | 0 | 0 | 0 | 0 | 0 | 1 | 0 | 1 | 0 | 0 |
| | 3 | 16 | 0 | 0 | 0 | 0 | 1 | 0 | 1 | 0 | 0 | 0 | 0 | 0 | 0 | 0 | 0 | 1 | 0 | 0 | 0 | 1 |
| | 4 | 17 | 0 | 0 | 0 | 0 | 1 | 1 | 0 | 0 | 0 | 0 | 0 | 0 | 0 | 0 | 0 | 1 | 0 | 0 | 1 | 0 |
| | 5 | 18 | 0 | 0 | 0 | 0 | 1 | 1 | 0 | 0 | 0 | 0 | 0 | 0 | 0 | 0 | 0 | 1 | 0 | 1 | 0 | 0 |
| | 6 | 19 | 0 | 0 | 0 | 0 | 1 | 1 | 0 | 0 | 0 | 0 | 0 | 0 | 0 | 0 | 0 | 1 | 0 | 0 | 0 | 1 |
| 8 | 1 | 20 | 0 | 1 | 0 | 0 | 0 | 0 | 1 | 0 | 0 | 0 | 0 | 0 | 0 | 0 | 0 | 1 | 0 | 0 | 0 | 0 |
| | 2 | 21 | 0 | 1 | 0 | 0 | 0 | 1 | 0 | 0 | 0 | 0 | 0 | 0 | 0 | 0 | 0 | 1 | 0 | 0 | 0 | 0 |
| 9 | 1 | 22 | 0 | 1 | 0 | 0 | 0 | 0 | 1 | 0 | 0 | 0 | 0 | 0 | 0 | 0 | 0 | 1 | 0 | 0 | 1 | 0 |
| | 2 | 23 | 0 | 1 | 0 | 0 | 0 | 1 | 0 | 0 | 0 | 0 | 0 | 0 | 0 | 0 | 0 | 1 | 0 | 0 | 1 | 0 |
| | 3 | 24 | 0 | 1 | 0 | 0 | 0 | 0 | 1 | 0 | 0 | 0 | 0 | 0 | 0 | 0 | 0 | 1 | 0 | 1 | 0 | 0 |
| | 4 | 25 | 0 | 1 | 0 | 0 | 0 | 1 | 0 | 0 | 0 | 0 | 0 | 0 | 0 | 0 | 0 | 1 | 0 | 1 | 0 | 0 |
| | 5 | 26 | 0 | 1 | 0 | 0 | 0 | 0 | 1 | 0 | 0 | 0 | 0 | 0 | 0 | 0 | 0 | 1 | 0 | 0 | 0 | 1 |
| | 6 | 27 | 0 | 1 | 0 | 0 | 0 | 1 | 0 | 0 | 0 | 0 | 0 | 0 | 0 | 0 | 0 | 1 | 0 | 0 | 0 | 1 |

**Table 6.** Continued

| Part *(k)* | Process route set *(PR(k))* | Process route number *(i)* | Machine no. (*m*) 1 | 2 | 3 | 4 | 5 | 6 | 7 | 8 | 9 | 10 | 11 | 12 | 13 | 14 | 15 | 16 | 17 | 18 | 19 | 20 |
|---|---|---|---|---|---|---|---|---|---|---|---|---|---|---|---|---|---|---|---|---|---|---|
| 10 | 1 | 28 | 0 | 1 | 0 | 0 | 1 | 1 | 0 | 0 | 0 | 0 | 0 | 0 | 0 | 0 | 0 | 1 | 0 | 0 | 0 | 0 |
| | 2 | 29 | 0 | 1 | 0 | 0 | 1 | 0 | 1 | 0 | 0 | 0 | 0 | 0 | 0 | 0 | 0 | 1 | 0 | 0 | 0 | 0 |
| 11 | 1 | 30 | 0 | 0 | 1 | 0 | 0 | 0 | 0 | 1 | 0 | 0 | 1 | 0 | 0 | 0 | 0 | 0 | 0 | 1 | 0 | 0 |
| | 2 | 31 | 0 | 0 | 1 | 0 | 0 | 0 | 0 | 1 | 0 | 0 | 1 | 0 | 0 | 0 | 0 | 0 | 0 | 0 | 1 | 0 |
| | 3 | 32 | 0 | 0 | 1 | 0 | 0 | 0 | 0 | 1 | 0 | 0 | 1 | 0 | 0 | 0 | 0 | 0 | 0 | 0 | 0 | 1 |
| 12 | 1 | 33 | 0 | 0 | 1 | 0 | 0 | 0 | 0 | 1 | 0 | 0 | 0 | 0 | 0 | 0 | 0 | 0 | 0 | 1 | 0 | 0 |
| | 2 | 34 | 0 | 0 | 1 | 0 | 0 | 0 | 0 | 1 | 0 | 0 | 0 | 0 | 0 | 0 | 0 | 0 | 0 | 0 | 1 | 0 |
| | 3 | 35 | 0 | 0 | 1 | 0 | 0 | 0 | 0 | 1 | 0 | 0 | 0 | 0 | 0 | 0 | 0 | 0 | 0 | 0 | 0 | 1 |
| 13 | 1 | 36 | 0 | 0 | 1 | 0 | 0 | 0 | 0 | 1 | 0 | 0 | 1 | 0 | 0 | 0 | 0 | 0 | 0 | 1 | 0 | 0 |
| | 2 | 37 | 0 | 0 | 1 | 0 | 0 | 0 | 0 | 1 | 0 | 0 | 1 | 0 | 0 | 0 | 0 | 0 | 0 | 0 | 1 | 0 |
| | 3 | 38 | 0 | 0 | 1 | 0 | 0 | 0 | 0 | 1 | 0 | 0 | 1 | 0 | 0 | 0 | 0 | 0 | 0 | 0 | 0 | 1 |
| 14 | 1 | 39 | 0 | 0 | 0 | 0 | 0 | 0 | 0 | 0 | 0 | 1 | 0 | 0 | 0 | 1 | 0 | 0 | 1 | 1 | 0 | 0 |
| | 2 | 40 | 0 | 0 | 0 | 0 | 0 | 0 | 0 | 0 | 0 | 1 | 0 | 0 | 0 | 1 | 0 | 0 | 1 | 0 | 1 | 0 |
| | 3 | 41 | 0 | 0 | 0 | 0 | 0 | 0 | 0 | 0 | 0 | 1 | 0 | 0 | 0 | 1 | 0 | 0 | 1 | 0 | 0 | 1 |
| 15 | 1 | 42 | 0 | 0 | 0 | 0 | 0 | 0 | 0 | 0 | 0 | 1 | 0 | 0 | 0 | 0 | 0 | 0 | 1 | 1 | 0 | 0 |
| | 2 | 43 | 0 | 0 | 0 | 0 | 0 | 0 | 0 | 0 | 0 | 1 | 0 | 0 | 0 | 0 | 0 | 0 | 1 | 0 | 1 | 0 |
| | 3 | 44 | 0 | 0 | 0 | 0 | 0 | 0 | 0 | 0 | 0 | 1 | 0 | 0 | 0 | 0 | 0 | 0 | 1 | 0 | 0 | 1 |
| 16 | 1 | 45 | 0 | 0 | 0 | 0 | 0 | 0 | 0 | 0 | 0 | 1 | 0 | 0 | 0 | 1 | 0 | 0 | 0 | 1 | 0 | 0 |
| | 2 | 46 | 0 | 0 | 0 | 0 | 0 | 0 | 0 | 0 | 0 | 1 | 0 | 0 | 0 | 1 | 0 | 0 | 0 | 0 | 1 | 0 |
| | 3 | 47 | 0 | 0 | 0 | 0 | 0 | 0 | 0 | 0 | 0 | 1 | 0 | 0 | 0 | 1 | 0 | 0 | 0 | 0 | 0 | 1 |
| 17 | 1 | 48 | 0 | 0 | 0 | 0 | 0 | 0 | 0 | 0 | 0 | 1 | 0 | 0 | 0 | 1 | 0 | 0 | 1 | 0 | 0 | 0 |
| 18 | 1 | 49 | 0 | 0 | 0 | 1 | 0 | 0 | 0 | 0 | 0 | 0 | 0 | 0 | 1 | 0 | 1 | 0 | 0 | 0 | 0 | 0 |
| 19 | 1 | 50 | 0 | 0 | 0 | 1 | 0 | 0 | 0 | 0 | 0 | 0 | 0 | 0 | 1 | 0 | 1 | 0 | 0 | 0 | 0 | 0 |
| 20 | 1 | 51 | 0 | 0 | 0 | 1 | 0 | 0 | 0 | 0 | 0 | 0 | 0 | 0 | 0 | 0 | 1 | 0 | 0 | 0 | 0 | 0 |

The seven cycles are shown in Figures 5–11.

From the seven cycles shown in Figures 5–11, seven process route families can be identified as:

(1) Process route family 1 consisting of process routes 2, 6 and 11 corresponding to parts 1, 3 and 5 respectively.

(2) Process route family 2 consisting of process routes 30, 33 and 36 corresponding to parts 11, 12 and 13 respectively.

(3) Process route family 3 consisting of process routes 49, 50 and 51 corresponding to parts 18, 19 and 20 respectively.

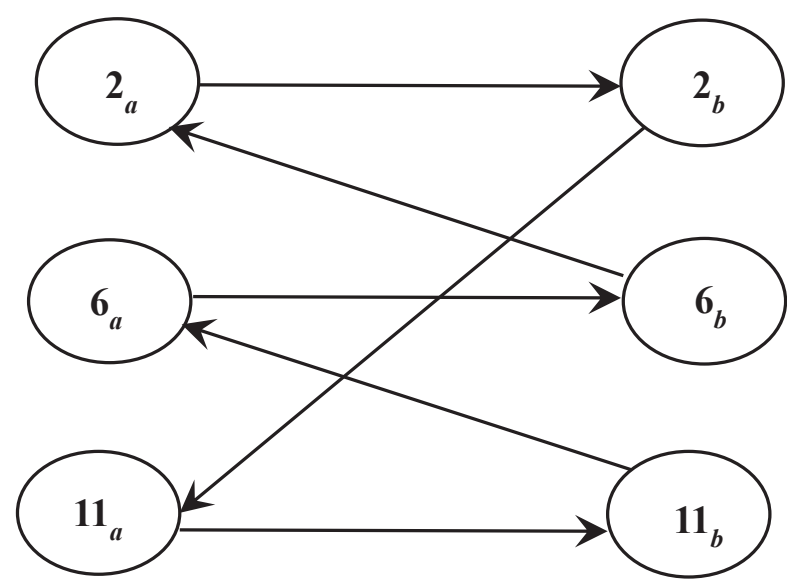

**Figure 5.** Cycle 1 for part family 2 of example problem II

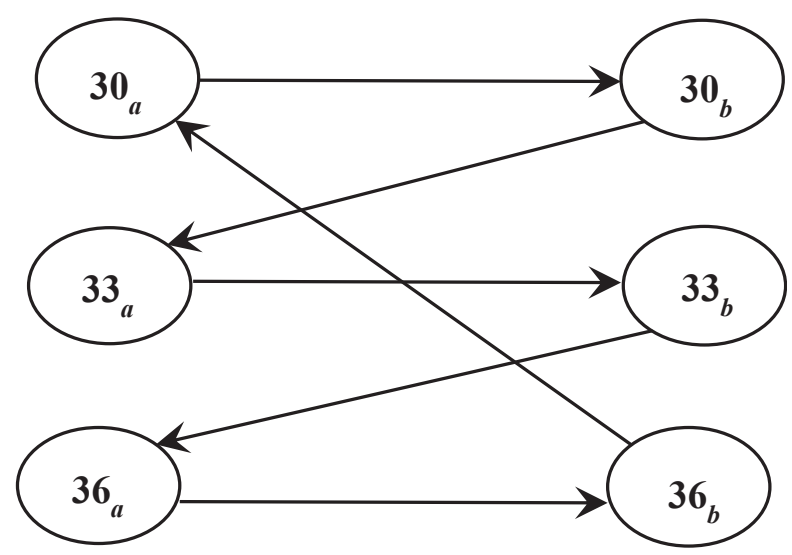

**Figure 6.** Cycle 2 for part family 2 of example problem II

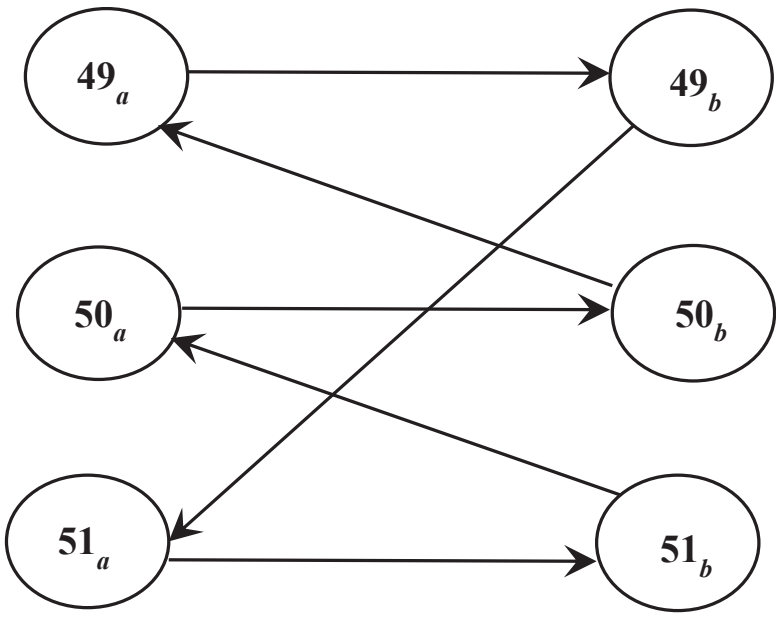

**Figure 7.** Cycle 3 for part family 2 of example problem II

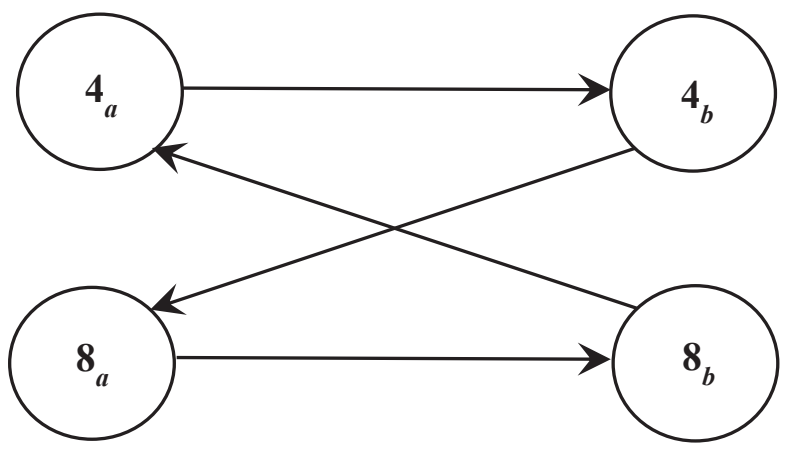

**Figure 8.** Cycle 4 for part family 2 of example problem II

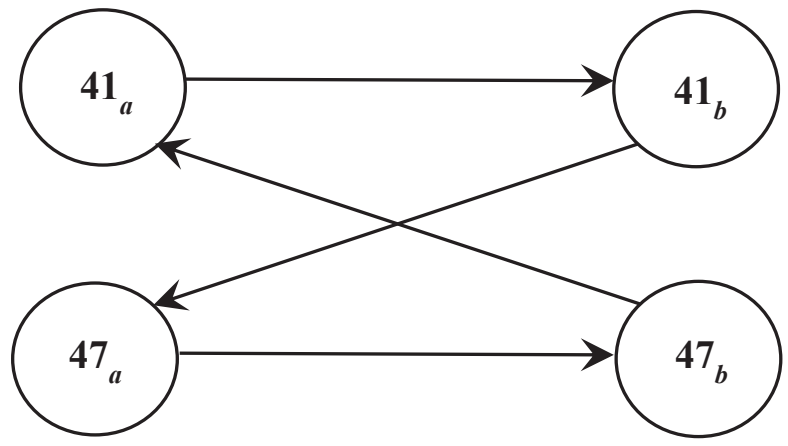

**Figure 9.** Cycle 5 for part family 2 of example problem II

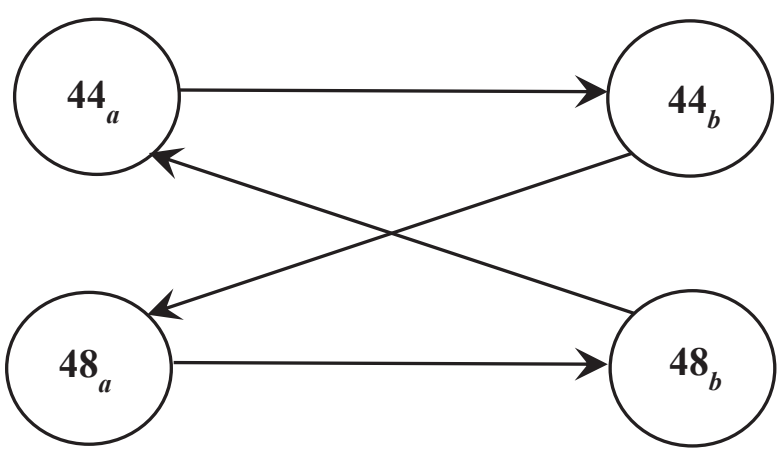

**Figure 10.** Cycle 6 for part family 2 of example problem II

(4) Process route family 4 consisting of process routes 4 and 8 corresponding to parts 2 and 4 respectively.

(5) Process route family 5 consisting of process routes 41 and 47 corresponding to parts 14 and 16 respectively.

(6) Process route family 6 consisting of process routes 44 and 48 corresponding to parts 15 and 17 respectively.

(7) Process route family 7 consisting of process routes 12, 17, 21, 23 and 28 corresponding to parts 6, 7, 8, 9 and 10 respectively.

In the machine cell formation stage, applying machine cell formation heuristic, five machine cells are obtained with only one exceptional element. These results are same as those obtained by solving the QAP formulation for machine cell formation:

(1) Machine cell 1 consisting of machines [1, 7, 9, 12]. Process route families as-signed to it are 1 and 4 which contain process routes [2, 4, 6, 8, 11].

(2) Machine cell 2 consisting of machines [2, 5, 6, 16, 19). Process route family assigned to it is 7 which contains process routes [12, 17, 21, 23, 28].

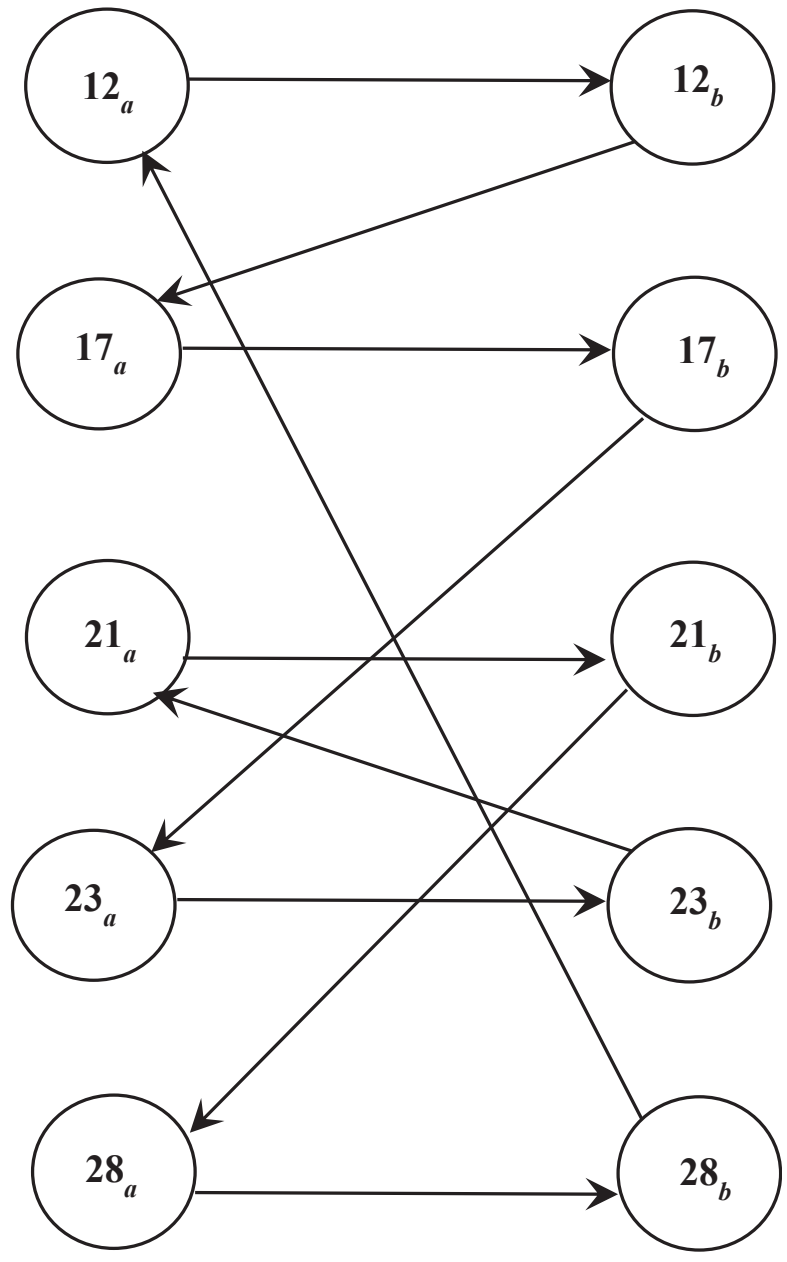

**Figure 11.** Cycle 7 for part family 2 of example problem II

(3) Machine cell 3 consisting of machines [3, 8, 11, 18]. Process route family assigned to it is 2 which contains process routes [30, 33, 36].

(4) Machine cell 4 consisting of machines [10, 14, 17, 20]. Process route families assigned to it are 5 and 6 which contain process routes [41, 44, 47, 48].

(5) Machine cell 5 consisting of machines [4, 13, 15]. Process route family assigned to it is 3 which contains process routes [49, 50, 51].

After rearranging the process route-machine incidence matrix, the resultant matrix is shown in Table 7. The present model yields one exceptional element, whereas the *p*-median formulation yields three exceptional elements. For comparison, the results obtained from the *p*-median model are shown in matrix form in Table 8.

*7.3 Practical relevance and industrial applicability*

While the two example problems presented in Sections 7.1 and 7.2 are drawn from established benchmark datasets in the generalized grouping literature, they are representative of the scale and structure encountered in real manufacturing facilities. Example Problem II, with K = 20 parts, $N$ = 51 process routes and M = 20 machines, models a production environment comparable to a small-to-medium enterprise (SME) manufacturing facility producing parts such as automobile spare components, precision machined parts or electronic sub-assemblies. Due to the availability of alternative machines with overlapping operational capabilities, which is a distinguishing feature of the generalized grouping problem, parts often have several possible process pathways in such systems.

To concretize the practical application of the proposed model, consider an automobile spare parts manufacturing facility that processes K = 25 distinct part types (e.g. shafts, gears, housings, brackets and flanges) across M = 15 machine types including CNC lathes, 3-axis and 5-axis

**Table 7.** Solution to example problem II in block diagonal form by present model

| Part (*k*) | Process route set (*PR* (*k*)) | Process route number (*i*) | Machine no. (*m*) | | | | | | | | | | | | | | | | | | | |
|---|---|---|---|---|---|---|---|---|---|---|---|---|---|---|---|---|---|---|---|---|---|---|
| | | | 1 | 9 | 12 | 7 | 2 | 5 | 6 | 16 | 19 | 3 | 8 | 11 | 18 | 10 | 14 | 17 | 20 | 4 | 13 | 15 |
| 1 | 2 | 2 | 0 | 1 | 1 | 1 | 0 | 0 | 0 | 0 | 0 | 0 | 0 | 0 | 0 | 0 | 0 | 0 | 0 | 0 | 0 | 0 |
| 2 | 2 | 6 | 1 | 1 | 1 | 1 | 0 | 0 | 0 | 0 | 0 | 0 | 0 | 0 | 0 | 0 | 0 | 0 | 0 | 0 | 0 | 0 |
| 3 | 3 | 11 | 1 | 1 | 1 | 0 | 0 | 0 | 0 | 0 | 0 | 0 | 0 | 0 | 0 | 0 | 0 | 0 | 1 | 0 | 0 | 0 |
| 4 | 2 | 4 | 1 | 0 | 1 | 1 | 0 | 0 | 0 | 0 | 0 | 0 | 0 | 0 | 0 | 0 | 0 | 0 | 0 | 0 | 0 | 0 |
| 5 | 2 | 8 | 1 | 0 | 1 | 1 | 0 | 0 | 0 | 0 | 0 | 0 | 0 | 0 | 0 | 0 | 0 | 0 | 0 | 0 | 0 | 0 |
| 6 | 1 | 12 | 0 | 0 | 0 | 0 | 1 | 1 | 1 | 0 | 0 | 0 | 0 | 0 | 0 | 0 | 0 | 0 | 0 | 0 | 0 | 0 |
| 7 | 4 | 17 | 0 | 0 | 0 | 0 | 0 | 1 | 1 | 1 | 1 | 0 | 0 | 0 | 0 | 0 | 0 | 0 | 0 | 0 | 0 | 0 |
| 8 | 2 | 21 | 0 | 0 | 0 | 0 | 1 | 0 | 1 | 1 | 0 | 0 | 0 | 0 | 0 | 0 | 0 | 0 | 0 | 0 | 0 | 0 |
| 9 | 2 | 23 | 0 | 0 | 0 | 0 | 1 | 0 | 1 | 1 | 1 | 0 | 0 | 0 | 0 | 0 | 0 | 0 | 0 | 0 | 0 | 0 |
| 10 | 1 | 28 | 0 | 0 | 0 | 0 | 1 | 1 | 1 | 1 | 0 | 0 | 0 | 0 | 0 | 0 | 0 | 0 | 0 | 0 | 0 | 0 |
| 11 | 1 | 30 | 0 | 0 | 0 | 0 | 0 | 0 | 0 | 0 | 0 | 1 | 1 | 1 | 1 | 0 | 0 | 0 | 0 | 0 | 0 | 0 |
| 12 | 1 | 33 | 0 | 0 | 0 | 0 | 0 | 0 | 0 | 0 | 0 | 1 | 1 | 0 | 1 | 0 | 0 | 0 | 0 | 0 | 0 | 0 |
| 13 | 1 | 36 | 0 | 0 | 0 | 0 | 0 | 0 | 0 | 0 | 0 | 1 | 1 | 0 | 1 | 0 | 0 | 0 | 0 | 0 | 0 | 0 |
| 14 | 3 | 41 | 0 | 0 | 0 | 0 | 0 | 0 | 0 | 0 | 0 | 0 | 0 | 0 | 0 | 1 | 1 | 1 | 1 | 0 | 0 | 0 |
| 15 | 3 | 47 | 0 | 0 | 0 | 0 | 0 | 0 | 0 | 0 | 0 | 0 | 0 | 0 | 0 | 1 | 1 | 0 | 1 | 0 | 0 | 0 |
| 16 | 3 | 44 | 0 | 0 | 0 | 0 | 0 | 0 | 0 | 0 | 0 | 0 | 0 | 0 | 0 | 1 | 0 | 1 | 1 | 0 | 0 | 0 |
| 17 | 1 | 48 | 0 | 0 | 0 | 0 | 0 | 0 | 0 | 0 | 0 | 0 | 0 | 0 | 0 | 1 | 1 | 1 | 0 | 0 | 0 | 0 |
| 18 | 1 | 49 | 0 | 0 | 0 | 0 | 0 | 0 | 0 | 0 | 0 | 0 | 0 | 0 | 0 | 0 | 0 | 0 | 0 | 1 | 1 | 1 |
| 19 | 1 | 50 | 0 | 0 | 0 | 0 | 0 | 0 | 0 | 0 | 0 | 0 | 0 | 0 | 0 | 0 | 0 | 0 | 0 | 1 | 1 | 1 |
| 20 | 1 | 51 | 0 | 0 | 0 | 0 | 0 | 0 | 0 | 0 | 0 | 0 | 0 | 0 | 0 | 0 | 0 | 0 | 0 | 1 | 1 | 1 |

**Table 8.** Solution to example problem II in block diagonal form by *p*-median model

| Part *(k)* | Process route set *(PR (k))* | Process route number *(i)* | Machine no. (*m*) | | | | | | | | | | | | | | | | | | | |
|---|---|---|---|---|---|---|---|---|---|---|---|---|---|---|---|---|---|---|---|---|---|---|
| | | | 7 | 9 | 12 | 1 | 20 | 2 | 5 | 6 | 16 | 19 | 3 | 8 | 11 | 18 | 10 | 14 | 17 | 4 | 13 | 15 |
| 1 | 2 | 2 | 1 | 1 | 1 | 0 | 0 | 0 | 0 | 0 | 0 | 0 | 0 | 0 | 0 | 0 | 0 | 0 | 0 | 0 | 0 | 0 |
| 2 | 2 | 4 | 1 | 0 | 1 | 1 | 0 | 0 | 0 | 0 | 0 | 0 | 0 | 0 | 0 | 0 | 0 | 0 | 0 | 0 | 0 | 0 |
| 3 | 2 | 6 | 1 | 1 | 1 | 1 | 0 | 0 | 0 | 0 | 0 | 0 | 0 | 0 | 0 | 0 | 0 | 0 | 0 | 0 | 0 | 0 |
| 4 | 2 | 8 | 1 | 0 | 1 | 1 | 0 | 0 | 0 | 0 | 0 | 0 | 0 | 0 | 0 | 0 | 0 | 0 | 0 | 0 | 0 | 0 |
| 5 | 3 | 11 | 0 | 1 | 1 | 1 | 1 | 0 | 0 | 0 | 0 | 0 | 0 | 0 | 0 | 0 | 0 | 0 | 0 | 0 | 0 | 0 |
| 6 | 1 | 12 | 0 | 0 | 0 | 0 | 0 | 1 | 1 | 1 | 0 | 0 | 0 | 0 | 0 | 0 | 0 | 0 | 0 | 0 | 0 | 0 |
| 7 | 4 | 17 | 0 | 0 | 0 | 0 | 0 | 0 | 1 | 1 | 1 | 1 | 0 | 0 | 0 | 0 | 0 | 0 | 0 | 0 | 0 | 0 |
| 8 | 2 | 21 | 0 | 0 | 0 | 0 | 0 | 1 | 0 | 1 | 1 | 0 | 0 | 0 | 0 | 0 | 0 | 0 | 0 | 0 | 0 | 0 |
| 9 | 2 | 23 | 0 | 0 | 0 | 0 | 0 | 1 | 0 | 1 | 1 | 1 | 0 | 0 | 0 | 0 | 0 | 0 | 0 | 0 | 0 | 0 |
| 10 | 1 | 28 | 0 | 0 | 0 | 0 | 0 | 1 | 1 | 1 | 1 | 0 | 0 | 0 | 0 | 0 | 0 | 0 | 0 | 0 | 0 | 0 |
| 11 | 1 | 30 | 0 | 0 | 0 | 0 | 0 | 0 | 0 | 0 | 0 | 0 | 1 | 1 | 1 | 1 | 0 | 0 | 0 | 0 | 0 | 0 |
| 12 | 1 | 33 | 0 | 0 | 0 | 0 | 0 | 0 | 0 | 0 | 0 | 0 | 1 | 1 | 0 | 1 | 0 | 0 | 0 | 0 | 0 | 0 |
| 13 | 1 | 36 | 0 | 0 | 0 | 0 | 0 | 0 | 0 | 0 | 0 | 0 | 1 | 1 | 1 | 1 | 0 | 0 | 0 | 0 | 0 | 0 |
| 14 | 2 | 40 | 0 | 0 | 0 | 0 | 0 | 0 | 0 | 0 | 0 | 1 | 0 | 0 | 0 | 0 | 1 | 1 | 1 | 0 | 0 | 0 |
| 15 | 2 | 43 | 0 | 0 | 0 | 0 | 0 | 0 | 0 | 0 | 0 | 1 | 0 | 0 | 0 | 0 | 1 | 0 | 1 | 0 | 0 | 0 |
| 16 | 2 | 46 | 0 | 0 | 0 | 0 | 0 | 0 | 0 | 0 | 0 | 1 | 0 | 0 | 0 | 0 | 1 | 1 | 0 | 0 | 0 | 0 |
| 17 | 1 | 48 | 0 | 0 | 0 | 0 | 0 | 0 | 0 | 0 | 0 | 0 | 0 | 0 | 0 | 0 | 1 | 1 | 1 | 0 | 0 | 0 |
| 18 | 1 | 49 | 0 | 0 | 0 | 0 | 0 | 0 | 0 | 0 | 0 | 0 | 0 | 0 | 0 | 0 | 0 | 0 | 0 | 1 | 1 | 1 |
| 19 | 1 | 50 | 0 | 0 | 0 | 0 | 0 | 0 | 0 | 0 | 0 | 0 | 0 | 0 | 0 | 0 | 0 | 0 | 0 | 1 | 1 | 1 |
| 20 | 1 | 51 | 0 | 0 | 0 | 0 | 0 | 0 | 0 | 0 | 0 | 0 | 0 | 0 | 0 | 0 | 0 | 0 | 0 | 1 | 0 | 1 |

CNC milling machines, drilling centers, cylindrical grinders, surface grinders and heat treatment furnaces. A single part type such as a "drive shaft" may have three feasible process routes: Route A uses a CNC lathe for turning followed by a cylindrical grinder, Route B uses a multi-axis turning center that integrates turning and grinding in a single setup and Route C uses a standard lathe followed by a different grinding machine. The process route–machine incidence matrix for this facility can be directly constructed from the shop floor routing sheets and the proposed network flow model can be applied without modification to determine: (1) the optimal process route selection for each part, (2) the resulting process route families based on minimum dissimilarity and (3) the machine cell configuration. The key practical advantage is that the model automatically determines the optimal number of process route families and does not require this number to be pre-specified, which eliminates a common source of suboptimality in practice.

Recent empirical studies confirm that mathematical cell formation methods translate effectively to industrial practice. Dhayef *et al.* (2024) applied cell formation methods at a state-owned electrical and electronic manufacturing company (Hydraulic Industries State Company, Baghdad, Iraq) with 10 machine types and 15 part types, demonstrating measurable improvements in grouping efficacy, system utilization and system flexibility relative to the facility's existing layout. Notably, their results showed that the optimized cellular layout reduced inter-cellular material handling movements by approximately 30% compared to the functional layout. Petrini and Sagawa (2026) developed and validated a production flow analysis approach for cell formation in a real manufacturing facility with multifunctional and universal machines, addressing practical complications such as technological constraints and workload balancing that arise in actual production environments. Their study specifically noted that the standard part–machine incidence matrix formulation (which is precisely the input format required by our model) can be effectively applied in real environments when combined with appropriate pre-processing of production flow data.

To provide a projection of the model's applicability to larger industrial settings, we note that a medium-sized automotive components factory might involve K = 50–100 part types with $N$ = 200–500 total process routes and M = 30–50 machines. As discussed in Section 5.4, the proposed model can handle instances of this scale within reasonable computation times using CPLEX and the heuristic procedure (Section 6.2) provides a computationally lightweight alternative that produced optimal results in our experiments. Future work will pursue two directions to further strengthen industrial validation: (1) applying the proposed model to large-scale, real-world datasets obtained from automotive and aerospace manufacturing facilities through industry collaboration and (2) integrating the model output with digital twin and cognitive digital twin frameworks, where the optimal cell configurations can serve as inputs for real-time simulation and adaptive shop-floor reconfiguration.

## 8. Conclusions and future research scopes

In this paper, we propose a procedure for forming part families and machine cells in a generalized grouping environment, where each part has more than one process route. The grouping problem involves selecting a process route for each part, grouping them into part families and forming machine cells such that each machine cell can process at least one process route (part) family. The procedure organizes the formation of part families and machine cells hierarchically. For the process route family formation stage, a unit capacity minimum cost network flow model is developed. The proposed model solves the part family formation problem optimally without pre-specifying the number of part families to be formed and is not an iterative process. The method is flexible, as it can be applied effectively in both scenarios: when disjoint groups exist and where they do not. Our computational results show that it provides better solutions than the *p*-median model in terms of exceptional elements.

For the machine cell formation stage, a QAP formulation and a heuristic procedure are proposed. The QAP formulation simultaneously assigns process route families and machines to the pre-specified number of cells to maximize machine utilization. A notable advantage of this formulation is that, even when a large number of cells is specified, the model autonomously determines the optimal number of cells within the specified limit, ensuring maximized machine utilization while satisfying system constraints. The heuristic procedure is hierarchical in nature. In the first stage, it aims to reduce the number of process route families wherever possible. In the second stage, machines are assigned to process route families based on maximum utilization among competing families. In the final stage, the procedure attempts to merge cells (and corresponding process route families) wherever feasible within system constraints. Computational results show that the QAP and heuristic procedure yield the same results.

In the presence of exceptional elements in the final solution, an improvement scheme can be employed. In reality, particularly in flexible manufacturing systems environments where machines have inherent flexibility, it may be possible to reassign exceptional operations that create inter-cell movements within the cell itself, provided that the reassignment does not violate machine capacity constraints. For solving the network model, we used a mixed-integer optimization package (CPLEX 22.1.0.0). Except for side constraint 7, all other constraints and the objective function conform to the standard unit capacity minimum cost network flow problem, which can be solved using any minimum cost network flow algorithm.

Due to constraint 7, a specialized network flow algorithm is required to solve the problem and the development of such an algorithm could be a future avenue of research. The network flow model developed in this work involves certain side constraints that prevent it from being solved using standard network flow algorithms. It would be worthwhile to develop a specialized network flow algorithm for this model.

It should also be acknowledged that the current model assumes deterministic process routes and machine requirements, which is a reasonable simplification for the foundational development of the network flow formulation. However, in dynamic manufacturing environments, process routes may be subject to stochastic variations due to machine breakdowns, variable processing times, fluctuating demand and other uncertainties. Extending the proposed model to accommodate such stochastic variables represents an important direction for future research. Possible approaches include robust optimization formulations that account for worst-case uncertainty in process route availability, stochastic programming extensions that incorporate probabilistic machine failure rates and integration with real-time monitoring systems (e.g. through cognitive digital twin frameworks) that can trigger re-optimization of process route families and cell configurations as manufacturing conditions change. Such extensions would significantly enhance the model's applicability to real-time, adaptive manufacturing systems.

**Authors' contributions**

Md. Kutub Uddin: Conceptualization, methodology development and manuscript writing. Md. Saiful Islam: Data analysis, validation and critical revisions. Md Abrar Jahin: Integer programming formulations, data analysis and manuscript writing. Md. Saiful Islam Seam: Literature review, results interpretation and manuscript writing. M.F. Mridha: Final review and project administration.

*Consent for publication*

All authors have read and approved the final manuscript and consent to its publication.

**Availability of data and material**

The datasets and models generated or analyzed during the study are available from the corresponding author upon reasonable request.

**Corresponding author**
Md Abrar Jahin can be contacted at: jahin@usc.edu